\documentclass[letterpaper, 10pt, conference]{ieeeconf}
\IEEEoverridecommandlockouts
\usepackage{cite}
\usepackage{amsmath,amssymb,amsfonts}
\usepackage{algorithmic}
\usepackage{graphicx}
\usepackage{textcomp}
\usepackage{xcolor}
\usepackage{siunitx}
\usepackage{float}

\renewcommand{\baselinestretch}{0.922}

\def\BibTeX{{\rm B\kern-.05em{\sc i\kern-.025em b}\kern-.08em
    T\kern-.1667em\lower.7ex\hbox{E}\kern-.125emX}}

\begin{document}
\title{\LARGE \bf A New $\boldsymbol{1}$-mg Fast Unimorph SMA-Based Actuator for Microrobotics\\
\thanks{This work was partially funded by the Washington State University (WSU) Foundation and the Palouse Club through a Cougar Cage Award to N. O. P\'erez-Arancibia. Additional funding was provided by the WSU Voiland College of Engineering and Architecture through a start-up fund to N. O. P\'erez-Arancibia.}% <-this % stops a space}
\thanks{C.~K.~Trygstad and N.~O.~Pérez-Arancibia are with the School of Mechanical and Materials Engineering, Washington State University (WSU), Pullman, WA 99164-2920, USA (e-mail: {\tt conor.trygstad@wsu.edu} (C.\,K.\,T.)); {\tt n.perezarancibia@wsu.edu} (N.\,O.\,P.-A.)).}
\thanks{X.-T.~Nguyen is with Flyby Robotics, Los Angeles, CA 90028, USA (e-mail: {\tt xuan-truc@flybydev.com}).}}
\author{Conor~K.~Trygstad, Xuan-Truc~Nguyen, and N\'estor~O.~P\'erez-Arancibia}
\maketitle
\thispagestyle{empty}
\pagestyle{empty}

\begin{abstract}
We present a new unimorph actuator for microrobotics, which is driven by thin \textit{shape-memory alloy} (SMA) wires. Using a \mbox{passive-capillary-alignment} technique and existing \mbox{SMA-microsystem} fabrication methods, we developed an actuator that is $\boldsymbol{7}$\,mm long, has a volume of $\boldsymbol{0.45}$\,mm\textsuperscript{$\boldsymbol{3}$}, weighs $\boldsymbol{0.96}$\,mg, and can achieve operation frequencies of up to $\boldsymbol{40}$\,Hz as well as lift $\boldsymbol{155}$ times its own weight. To demonstrate the capabilities of the proposed actuator, we created an \mbox{$\boldsymbol{8}$-mg} crawler, the \mbox{MiniBug}, and a bioinspired \mbox{$\boldsymbol{56}$-mg} controllable \mbox{water-surface-tension} crawler, the \mbox{WaterStrider}. The \mbox{MiniBug} is $\boldsymbol{8.5}$\,mm long, can locomote at speeds as high as $\boldsymbol{0.76}$\,BL/s (\textit{\mbox{body-lengths~per~second}}), and is the lightest \mbox{fully-functional} crawling microrobot of its type ever created. The WaterStrider is $\boldsymbol{22}$\,mm long, and can locomote at speeds of up to $\boldsymbol{0.28}$\,BL/s as well as execute turning maneuvers at angular rates on the order of $\boldsymbol{0.144}$\,rad/s. The \mbox{WaterStrider} is the lightest controllable \mbox{SMA-driven} \mbox{water-surface-tension} crawler developed to date. 
\end{abstract}

\section{Introduction}
\label{SECTION01}
We envision the creation of autonomous \mbox{millimeter-scale} robots that can work together in swarms to execute tasks such as artificial pollination, \mbox{insect-plague} control, agricultural surveying, search and rescue, environmental monitoring, microfabrication, and \mbox{robotic-assisted} surgeries. In order for \mbox{millimeter-scale} robots to complete complex assignments and function autonomously, they must efficiently generate large forces relative to their size and weight. The most common actuators used in microrobotics are based on piezoelectric~\cite{lee2011design,goldberg2018power,wu2019insect,gravish2020stcrawler}, electromagnetic~\cite{hollar2003solar,contreras2017first,vogtmann201725,lu2018bioinspired,hu2018small,pierre20183d}, or \mbox{\textit{shape-memory alloy}} (SMA)~\cite{yang2007new,kim2019laser,SMALLBug,SMARTI,yang202088} technologies. Because of their wider frequency bandwidths, piezoelectric and electromagnetic microactuators are generally preferred over \mbox{SMA-based} systems. However, compared to \mbox{SMA-based} methods, these two actuation technologies exhibit significantly lower work densities and, additionally, electromagnetic microactuators have been shown to be difficult to operate outside laboratory settings. In~\cite{SMALLBug}, we introduced a $6$-mg fast \mbox{SMA-based} microactuator; here, we present an actuator of the same type which is significantly lighter, smaller, and stronger for its size. In fact, this is the lightest and fastest \mbox{SMA-based} actuator reported to date; it weighs only $0.96$\,mg, has a length of $7$\,mm, has a volume of $0.45$\,mm$^3$, can produce functional displacements at frequencies of up to $40$\,Hz, and can lift $155$~times its own weight. This achieved performance is a consequence of the actuator's mechanical design and structure, which is essentially composed of two parallel nitinol ($56$\,\% Nickel~\&~$44$\,\% Titanium) SMA wires with a diameter of \mbox{$25.4$\,{\textmu}m} and a leaf spring used to facilitate the transition between the \textit{twinned} and \textit{detwinned martensite} phases of the SMA material during an actuation cycle~\cite{SMALLBug}. This configuration is the key element that enables the actuator to operate at high frequencies (up to $40$\,Hz).
\begin{figure}[t!]
\vspace{1ex}
\begin{center}
\includegraphics{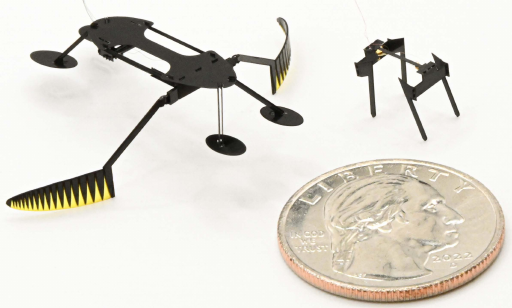}
\end{center}
\vspace{-2ex}
\caption{\textbf{Two microrobots driven by the proposed \mbox{SMA-based} actuator.} The WaterStrider (left) is a $56$-mg controllable robot with a body length of $22$\,mm that crawls on water by taking advantage of \mbox{surface-tension} phenomena. The MiniBug (right) is an \mbox{$8$-mg} microrobot with a body length of $8.5$\,mm that crawls on land. This robot is the lightest \mbox{fully-functional} \mbox{SMA-driven} terrestrial crawler developed to date. 
\label{FIG01}}
\vspace{-2ex}
\end{figure}

To test and demonstrate the capabilities of the proposed microactuator, we developed an \mbox{$8$-mg} terrestrial crawler, the MiniBug, and a bioinspired \mbox{$56$-mg} \mbox{water-surface-tension} crawler, the WaterStrider; these two robots are shown in Fig.\,\ref{FIG01}. Inspired by the design introduced in~\cite{SMALLBug}, the MiniBug is the lightest \mbox{fully-functional} \mbox{SMA-driven} crawler ever created and represents significant progress
toward miniaturization with respect to the crawling platforms in~\cite{SMALLBug}~and~\cite{SMARTI}. Specifically, the MiniBug has a length of \mbox{$8.5$\,mm} and can locomote at speeds of up to \mbox{$0.76$\,BL/s}~(\textit{body-lengths~per~second}). We envision that robots of this type can be equipped with micromanipulators and be used in swarms to execute tasks in structured environments with smooth surfaces; for example, to provide assistance in \mbox{small-scale} production lines. Inspired by the locomotion modes of common \mbox{water-strider} insects (\textit{Gerris lacustris}), the WaterStrider prototype (see Fig.\,\ref{FIG01}) has a lightweight structure and elliptical supporting feet with relatively large surface areas, which ensure that the generated \mbox{surface-tension} forces are strong enough to compensate gravity and allow the robot to stably stand on water. The two \mbox{fin-like} propulsors of this platform are independently driven by two of the proposed \mbox{high-work-density} actuators through \mbox{four-bar} transmissions designed and fabricated using the \textit{smart composite microstructure}~(SCM) method. During operation, the typical resulting stroke angles are on the order of $30\,^{\circ}$ and directional control of the robot is achieved by changing their amplitudes in real time (see Section\,\ref{SECTION04}). In this case, through simple experiments and heuristic considerations, we aimed to maximize the hydrodynamic force output by taking advantage of \mbox{fluid-structure-interaction} phenomena, a matter of current and further research. The WaterStrider prototype in Fig.\,\ref{FIG01} can locomote at speeds of up to \mbox{$0.28$\,BL/s} and complete \mbox{open-loop} turning maneuvers at angular rates of up to \mbox{$0.144$\,rad/s}. We anticipate that in the near future, \mbox{water-surface-tension} crawlers of the WaterStrider type will be employed for geographical surveying and continuous water quality monitoring in lakes, dams, and rivers; for example, to quickly detect toxic spills or changes in hydrology. 
 
The rest of the paper is organized as follows. Section\,\ref{SECTION02} describes the design and fabrication of the proposed unimorph \mbox{SMA-based} actuator. Section\,\ref{SECTION03} discusses the experimental characterization of the actuator. Section\,\ref{SECTION04} describes the development and functionality of the MiniBug and WaterStrider platforms, which we created to test and demonstrate the capabilities of the actuator. Lastly, Section\,\ref{SECTION05} draws some conclusions regarding the presented research.
\begin{figure}[t!]
\vspace{1ex}
\begin{center}
\includegraphics{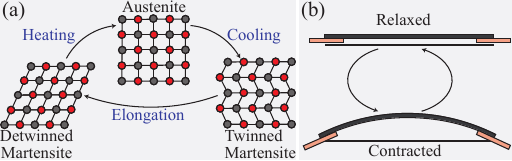}
\end{center}
\vspace{-2ex}
\caption{\textbf{Design and functionality of the proposed \mbox{SMA-based} actuator.} \textbf{(a)}~Depiction of the molecular crystal structure of an SMA material during cycles of heating and cooling, assuming the completion of major hysteretic loops. In the case of an SMA wire, starting at the elongated \textit{detwinned martensite} phase, the SMA material reaches the \textit{austenite} phase after the application of sufficient heat to surpass the SMA transition temperature and thus force the contraction of the wire according to the \textit{\mbox{shape-memory} effect} (SME). Then, after sufficient cooling, the material transitions to the \textit{twinned martensite} phase. As shown, the application of an external stress detwins the SMA material and the wire elongates until reaching its initial state. \textbf{(b)}~Depiction of a complete \mbox{SME-based} actuation cycle during operation. Heat is applied to the SMA wire using \textit{Joule} heating; cooling of the SMA material occurs passively through unforced convection; simultaneously, the SMA material is detwinned using a CF leaf spring. \label{FIG02}}
%\vspace{-1ex}
%\end{figure}
\vspace{2ex}
%\begin{figure}[t!]
%\vspace{1ex}
\begin{center}
\includegraphics{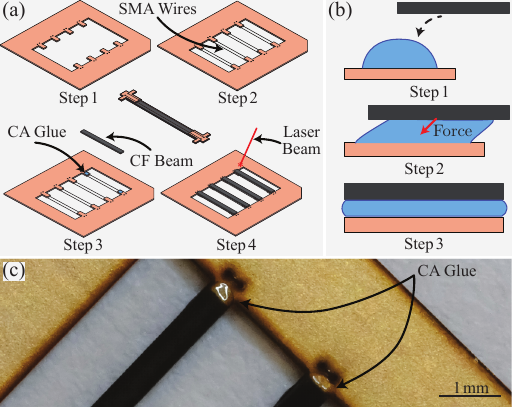}
\end{center}
\vspace{-2ex}
\caption{\textbf{Fabrication of the proposed SMA-based actuator.} \textbf{(a)}~Steps of the \mbox{SCM-based} fabrication procedure. In Step\,1, a \mbox{$3$-W} \mbox{$355$-nm} DPSS laser (Photonics~Industries~\mbox{DCH-355-3}) is used to micromachine the \mbox{Cu-FR4} supporting frame employed during fabrication. In Step\,2, SMA wires are looped through holes in the frame, and secured under tension with a simple knot and a small amount of CA glue (Loctite~401). In Step\,3, we use capillary \mbox{self-alignment} to accurately place four CF beams onto premachined alignment tabs of the \mbox{Cu-FR4} frame. In Step\,4, the actuator is released from the \mbox{Cu-FR4} frame using a \mbox{$3$-W} \mbox{$355$-nm} DPSS laser. \textbf{(b)}~Capillary alignment used in Step\,3 of the actuator fabrication process. In Step\,1, droplets of CA glue are precisely placed onto alignment tabs of the \mbox{Cu-FR4} frame, and four CF beams are immediately placed on top of the droplets. In Step\,2, each beam is pulled, without external intervention, in the desired direction of alignment due to the surface tension of the glue and capillary action at the \mbox{beam-glue} interface. In Step\,3, the CA glue cures and the CF beams become precisely aligned over the tabs of the \mbox{Cu-FR4} frame. \textbf{(c)}~Photo showing two cured CA glue droplets and two precisely aligned beams after completion of Step\,3 in the actuator fabrication process. \label{FIG03}}
\vspace{-2ex}
\end{figure}

\section{Design and Fabrication} 
\label{SECTION02}
The dynamic behavior of an SMA wire is characterized by the \textit{shape memory effect}~(SME) and the \textit{superelasticity} property~\cite{lagoudas2008shape}. These phenomena are observable when the SMA material composing the wire transitions between three distinct \mbox{crystal-structure} phases: \textit{detwinned martensite}, \textit{twinned martensite}, and \textit{austenite}. As shown in Fig.\,\ref{FIG02}(a), an actuation cycle can be induced through sequential heating, cooling, and application of stress. The proposed design can be driven using \textit{Joule} heating~\cite{joey2019position}, or other methods such as catalytic combustion~\cite{yang202088}, and passive cooling (free convection); stress is continuously applied using a leaf spring made of \textit{carbon fiber}~(CF), as shown in Fig.\,\ref{FIG02}(b). In this specific case, the SMA material (nitinol), under a stress of $172$\,MPa, has a nominal transition temperature from martensite to austenite of $90$\,\textdegree{C}. Using basic beam theory, it can be shown that the force applied by the leaf spring is approximately constant for contractions of the actuator's SMA wires greater than $18$\,{\textmu}m; therefore, for design purposes, we assumed that the stress experienced by the SMA material remains constant during an entire actuation cycle. As seen in Fig.\,\ref{FIG03}, from a fabrication viewpoint, the proposed \mbox{SMA-based} actuator is composed of three main types of components: (i)~two parallel \mbox{$25.4$-{\textmu}m-diameter} SMA wires (Dynalloy Flexinol HT SMA Wire); (ii)~a $90$-{\textmu}m-thick CF beam, made of Tenax 112 prepreg, that functions as a leaf spring; and, (iii)~two \mbox{cross-shaped} plates, made of \mbox{copper-clad}~FR4~(Cu-FR4), used for electrical and mechanical connection.

Using data reported in~\cite{SMALLBug} and simple mechanical tests, we chose a thickness of $90$\,{\textmu}m, a width of $0.5$\,mm, and a length of $6$\,mm for the CF beam in order to heuristically minimize weight and maximize actuator output at high frequencies. Similarly, we chose the smallest \mbox{commercially-available} diameter of $25.4$\,{\textmu}m for the SMA wires in order to minimize the volume with respect to the surface of the SMA material, thus maximizing the \mbox{free-convection} rate of cooling. In this case, the design \mbox{trade-off} is that the force produced by an SMA wire decreases with its diameter. To compensate, we can simply increase the number of SMA wires used for actuation; as already mentioned, the presented design uses two in parallel. All the steps in the simultaneous fabrication process of four actuators are detailed in Fig.\,\ref{FIG03}(a). In \mbox{Step\,1}, a \mbox{Cu-FR4} frame is cut using a \mbox{$3$-W} $355$-nm DPSS laser (Photonics Industries \mbox{DCH-355-3}). In \mbox{Step\,2}, SMA wires are looped through orifices at two opposite sides of the \mbox{Cu-FR4} frame and tied under tension using a simple knot; then, they are secured using a small amount of \textit{cyanoacrylate}~(CA) glue (Loctite\,401). In \mbox{Step\,3}, CF beams are installed on protrusions (alignment tabs) of the frame, using the passive capillary \mbox{self-alignment} phenomena described in~\cite{chang2016capillary,SurfaceTensionSelfAlignment}; at this scale, the properties of the chosen CA glue produce the capillary effects necessary for passive \mbox{self-alignment} during the short period of time before curing. 

The mechanism of the \mbox{capillary-alignment} technique is depicted in~Fig.\,\ref{FIG03}(b). First, droplets of CA glue are precisely deposited on the alignment tabs of the \mbox{Cu-FR4} frame; then, premachined CF beams are placed on top of the droplets; lastly, with proper droplet placement, the capillary forces of the CA glue \textit{pull} the CF beams perfectly over the \mbox{Cu-FR4} alignment tabs before the glue cures. Because of the quick cure time of CA glue, the CF beams must be placed immediately after droplet deposition to ensure proper capillary alignment. Specifically, the beams are passively aligned on the \mbox{Cu-FR4} tabs with respect to their transverse direction while slight manipulations with tweezers are required to center them with respect to their axial direction. In Step\,4, as depicted in Fig.\,\ref{FIG03}(a), the actuators are released using DPSS laser cutting. The photograph in~Fig.\,\ref{FIG03}(c) shows a \mbox{close-up} of an \mbox{actuator-fabrication} frame after the installation of CF beams; here, the precise alignment of two beams on their \mbox{Cu-FR4} tabs can be seen as well as the cured CA glue droplets holding them in place. 

\section{Actuator Characterization}
\label{SECTION03}
\subsection{Experimental Setup}
\begin{figure}[t!]
\vspace{1.4ex}
\begin{center}
\includegraphics{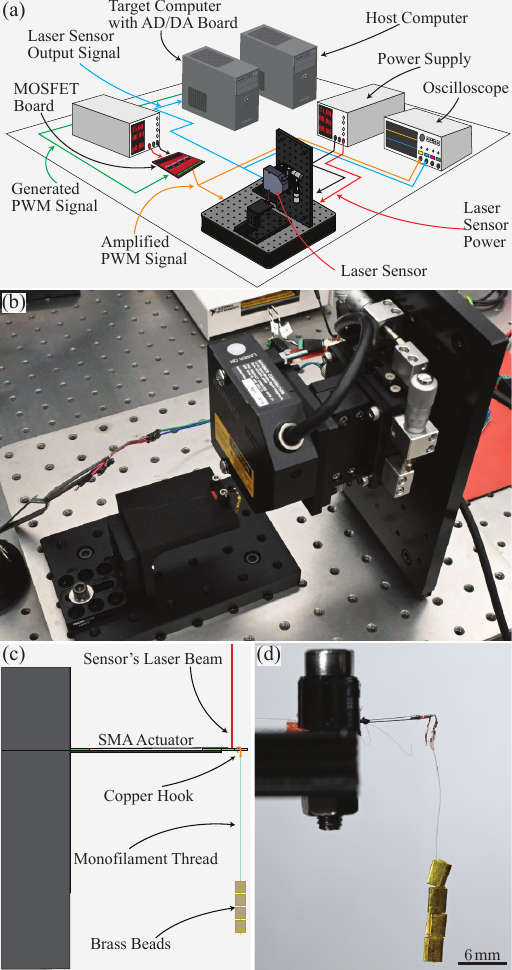}
\end{center}
\vspace{-0.8ex}
\caption{\textbf{Experimental setup used to dynamically characterize the proposed actuator}. \textbf{(a)}~\mbox{Signals-and-systems} diagram of the setup. A MathWorks \mbox{target-and-host} Simulink \mbox{Real-Time} system with a National Instruments \mbox{PCI-6229} \mbox{AD/DA} board is used to digitally generate signals, read and record measured variables, and process information at a rate of $10$\,kHz. Accordingly, the PWM signal required to excite the tested actuator is first generated and then its power is amplified with a \mbox{MOSFET-based} circuit (\mbox{YYNMOS-4}) to provide sufficient current while \textit{Joule} heating the SMA wires of the tested actuator. A laser displacement sensor (\mbox{Keyence~LK-G32}) 
measures the instantaneous deflection of the actuator's tip, the output. An oscilloscope monitors signals in real time and records the PWM data coming from the \mbox{MOSFET-based} circuit. \textbf{(b)}~Photograph of the test stand with an actuator mounted. \textbf{(c)}~Schematic of the \mbox{weight-loading} method and fixture used to hold an actuator during the characterization tests. The tested actuator is fixed on one end with the distal free end precisely aligned, using a \mbox{3-axis} optomechanical stage, under the laser displacement sensor shown in~(b). We use a copper hook to attach a piece of \mbox{monofilament} thread to the actuator; we secure it with a small amount of CA glue. We increase the loading acting on the actuator by crimping additional brass beads, weighing $0.18$\,mN each, to the thread. After they are crimped, the beads are secured using CA glue to prevent shaking during operation. To compensate for the $0.015$-mN weight of the hook and thread, the first crimped bead weighs only $0.165$\,mN. \textbf{(d)}~Photo of a tested actuator mounted on the experimental stand with a $0.72$\,mN load hanging from it. \label{FIG04}}
\vspace{-6ex}
\end{figure}

For the experimental characterization of the proposed actuator dynamics, we used the setup shown in~Fig.\,\ref{FIG04}. As seen in the \mbox{signals-and-systems} diagram of~Fig.\,\ref{FIG04}(a), a \mbox{MathWorks} Simulink \mbox{Real-Time} system is used to generate the \mbox{\textit{pulse-width-modulation}} (PWM) voltage signal with \mbox{pre-specified} characteristics\textemdash{frequency, \mbox{\textit{on}-voltage} height, and \textit{duty cycle} (DC)\textemdash}that drives the actuator. The power of this PWM signal is amplified with a \mbox{MOSFET-based} circuit (\mbox{YYNMOS-4}) to provide sufficient current to Joule heat the SMA wires during actuation. Throughout characterization, the tested actuator is mounted on the stand shown in~Fig.\,\ref{FIG04}(b) and depicted in~Fig.\,\ref{FIG04}(c). In this configuration, one end of the actuator is attached to a \mbox{3D-printed} mount while the other end is precisely aligned below a laser displacement sensor (\mbox{Keyence~LK-G32}) to measure the instantaneous displacement output of the actuator. During the tests, signals are digitally generated, measured variables are read and recorded, and information is processed at a rate of $10$\,kHz.
\begin{figure*}[t!]
\vspace{1ex}
\begin{center}
\includegraphics[width=1.0\textwidth]{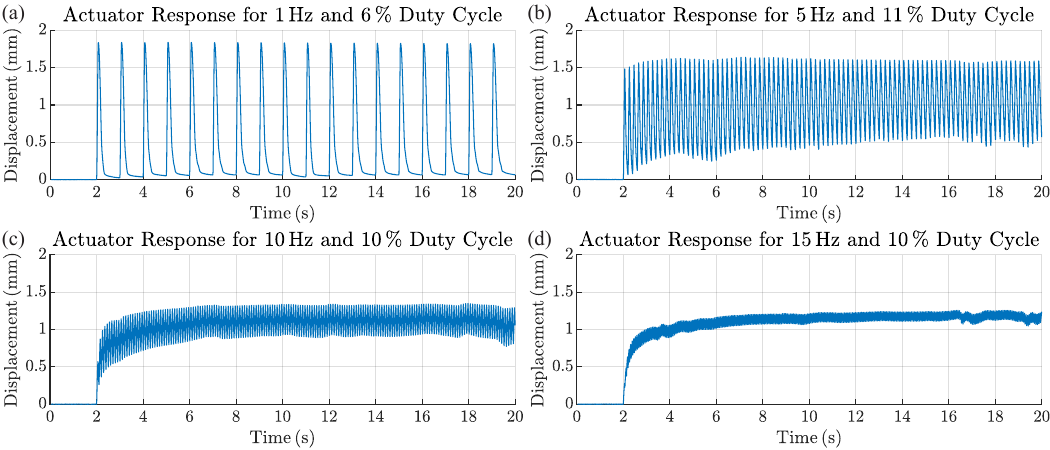}
\end{center}
\vspace{-2ex}
\caption{\textbf{Actuator responses measured with a laser displacement sensor.} \textbf{(a)}~Actuator response to a PWM signal with a frequency of $1$\,Hz and DC of $6$\,\%. \textbf{(b)}~Actuator response to a PWM signal with a frequency of $5$\,Hz and DC of $11$\,\%. \textbf{(c)}~Actuator response to PWM signal with a frequency of $10$\,Hz and DC of $10$\,\%. \textbf{(d)}~Actuator response to a PWM signal with a frequency of $15$\,Hz and DC of $10$\,\%. In these four cases, the \textit{on}-voltage height of the PWM signal is $15$\,V. \label{FIG05}}
\vspace{-2ex}
\end{figure*}

During one PWM period, the voltage applied across an SMA wire of the actuator is first \textit{on}, then \textit{off}; here, \textit{on} corresponds to $15$\,V and \textit{off} corresponds to $0$\,V. The contraction rate of an SMA wire directly depends on the amount of current running through it~\cite{joey2019position,SMALLBug}, and the DC of the driving PWM signal\textemdash{defined as the fraction of the signal period for which the signal is in the \textit{on} state\textemdash}determines the fraction of time an actuator is allowed to heat during a PWM cycle~\cite{SMALLBug}. At low operation frequencies ($\sim{\hspace{-0.6ex}}1$\,Hz), a large DC might result in the SMA material becoming overheated and damaged. Therefore, using information from simple tests, we selected a set of DC values for each considered frequency. We chose an \mbox{\textit{on}-voltage} height of $15$\,V to limit the current passing through the SMA wires of the actuator to $200$\;mA. This limitation is critical to avoid overheating and damaging the SMA material and thin copper wires ($52$\,AWG) used to connect the actuator to the driving power circuit. 

To empirically evaluate the performance of an actuator regarding force and work, we load and operate it with several different weights for several different PWM parameters. \mbox{Figs.\,\ref{FIG04}(b)--(d)} graphically describe the method to load the actuator. As seen, a short piece of \mbox{monofilament} thread is connected to the distal end of the actuator through a copper hook; then, brass beads weighing \mbox{$0.18$\,mN} each are incrementally crimped onto the thread until the actuator can no longer lift the load. To compensate for the combined \mbox{$0.015$-mN} weight of the hook and thread, the first bead was chosen to weigh only \mbox{$0.165$\,mN}. After data collection, the displacement data measured with the laser sensor is processed offline using a \mbox{zero-phase} \mbox{low-pass} filter\textemdash{designed with MATLAB's \mbox{digital-filter-design} tool\textemdash}to reduce sensor noise. We then employ simple algorithms run in MATLAB to compute figures of merit to evaluate the performance of the proposed actuator.

\subsection{Characterization Results} 
\begin{figure}[t!]
\vspace{1ex}
\begin{center}
\includegraphics{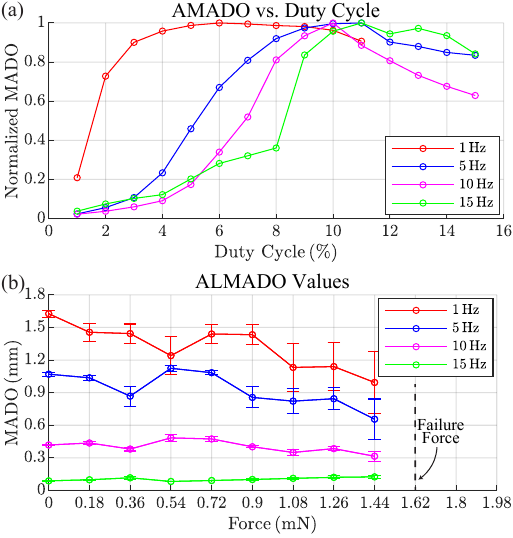}
\end{center}
\vspace{-2ex}
\caption{\textbf{Normalized \textit{average} MADO (AMADO) and \textit{average~loaded} MADO (ALMADO) values measured experimentally}. \textbf{(a)}~AMADO values corresponding to all sixty cases defined by the exciting PWM frequencies in the set $\left\{1,5,10,15\right\}$\,Hz and PWM DC values in the set $\left\{1,2,3,4,5,6,7,8,9,10,11,12,13,14,15 \right\}$\,\%. At $1$\,Hz, the largest AMADO value of $1.625$\,mm corresponds to a DC of $6$\,\%. At $5$\,Hz, the largest AMADO value of $1.15$\,mm corresponds to a DC of $11$\,\%. At $10$\,Hz, the largest AMADO value of $0.48$\,mm corresponds to a DC of $10$\,\%. At $15$\,Hz, the largest AMADO value of $0.14$\,mm corresponds to a DC of $10$\,\%. During the performance of these experiments, the measured ambient temperature oscillated approximately two degrees about $22\,^{\circ}$C. \textbf{(b)}~ALMADO data corresponding to all thirty-six cases defined by the exciting PWM frequency-DC pairs in the set $\left\{\left\{ 1\,\textrm{Hz},6\,\textrm{\%}\right\}, \left\{ 5\,\textrm{Hz},11\,\textrm{\%} \right\},
\left\{ 10\,\textrm{Hz},10\,\textrm{\%} \right\},
\left\{ 15\,\textrm{Hz},10\,\textrm{\%} \right\}
\right\}$ and the loads in the set $\left\{0,0.18,0.36,0.54,0.72,0.90,1.08,1.26,1.44 \right\}$\,mN. Each data point in the plot denotes the mean of the ALMADO values obtained through five different \mbox{back-to-back} experiments. The associated \textit{standard error of the mean} (SEM) values are indicated with vertical bars. For each tested load, the ALMADO value significantly decreases as the exciting frequency increases. Also, for the first three frequencies, a decreasing trend of the ALMADO value, as the load increases, can be observed. Namely, for the pair $\left\{1\,\textrm{Hz},6\,\textrm{\%}\right\}$, the mean of this figure of merit decreases from $1.625$ to $0.994$\,mm. Similarly, for the pair $\left\{5\,\textrm{Hz},11\,\textrm{\%}\right\}$, it decreases from $1.15$ to $0.655$\,mm; and, for the pair $\left\{10\,\textrm{Hz},10\,\textrm{\%}\right\}$, from $0.48$ to $0.315$\,mm. In contrast, for the pair $\left\{15\,\textrm{Hz},10\,\textrm{\%}\right\}$, the ALMADO value remains approximately constant as the load increases. \label{FIG06}}
\vspace{-2ex}
\end{figure}

Actuator responses for PWM exciting frequencies of \mbox{$1$, $5$, $10$, and $15$\,Hz} are shown in Figs.\,\ref{FIG05}(a)--(d), respectively. As indicated in the plots, the associated DC values are $6$, $11$, $10$, and $10$\,\%, which were empirically determined to maximize actuator output. Sections of these experiments can be seen in the accompanying supplementary movie. At $1$\,Hz, the measured \mbox{steady-state} deflection at the actuator's tip oscillates between about $0.1$ and $1.75$\,mm, which approximately corresponds to a major hysteretic loop~\cite{lagoudas2008shape}. At higher frequencies, the amplitude of oscillation decreases and a \mbox{steady-state} deflection offset occurs. These phenomena result from the hysteretic behavior of the SMA material during \mbox{heating-and-cooling} cycles, and the limited time available for the SMA wires to cool down and reach ambient temperature. To evaluate actuator performance, we define the \textit{maximum actuator displacement output} (MADO) for an actuation cycle as the difference between the maximum and minimum beam deflection at the actuator's tip during a PWM period, measured using the laser sensor shown in Fig.\,\ref{FIG04}(b). Furthermore, for a test defined by its PWM exciting frequency and PWM DC value, we define the \textit{average}~MADO (AMADO) as the mean of the MADO sequence for a test, computed across $15$\,s of \mbox{steady-state} data. Fig.\,\ref{FIG06}(a) shows the AMADO values corresponding to all sixty experimental cases defined by the exciting PWM frequencies in the set $\left\{1,5,10,15\right\}$\,Hz and PWM DC values in the set $\left\{1,2,3,4,5,6,7,8,9,10,11,12,13,14,15\right\}$\,\%. For plotting and analysis purposes, for each PWM frequency, we normalized the AMADO data by dividing them by the maximum computed AMADO value among all the tested DC values. As mentioned above, for the elements in the chosen frequency set, these maxima respectively occur when the DC values are chosen to be $6$, $11$, $10$, and $10$\,\%; correspondingly, the specific maximum raw AMADO values are $1.625$, $1.15$, $0.48$, and $0.14$\,mm. As seen, the AMADO value for an experiment significantly decreases as the exciting PWM frequency increases; however, at $15$\,Hz, the AMADO value of $140$\,{\textmu}m is still comparable to the displacement a piezoelectric actuator of this scale can produce~\cite{beePlus2019}. 

During \mbox{force-output} characterization experiments, for each PWM exciting frequency, we employ the DC value corresponding to the largest \mbox{load-free} AMADO value. In the cases presented here, for each load in the set $\left\{0,0.18,0.36,0.54,0.72,0.90,1.08,1.26,1.44\right\}$\,mN and each frequency in the set $\left\{1,5,10,15\right\}$\,Hz, we measured the instantaneous loaded deflection, $d_{\textrm{max}}(f_{\textrm{load}})$, of the actuator's tip and compute the \mbox{\textit{average~loaded}~MADO} ($\textrm{ALMADO}$) value across $15$\,s, $\bar{d}_{\textrm{max}}(f_{\textrm{load}})$, where $f_{\textrm{load}}$ is the corresponding load. The chosen load set is a reflection of the way we increment the weight of the test load in each sequential experiment. Specifically, in a first set of experiments, the actuator is excited unloaded; then, in a second set of experiments, we add a weight of \mbox{$0.18$\,mN}, corresponding to the hook, the monofilament thread, and a \mbox{$0.165$-mN} brass bead; then, in a third set of experiments, we crimp a \mbox{$0.18$-mN} bead to the thread, and so forth, as depicted in Figs.\,\ref{FIG04}(c)--(d). We determined empirically that eight brass beads (corresponding to \mbox{$1.44$\,mN}) can be added before an actuator of the considered type fails.

The mean and \textit{standard error of the mean} (SEM) of the ALMADO values obtained through five different experiments, for each considered frequency and DC value, are shown in Fig.\,\ref{FIG06}(b). Using these data, we can estimate the \textit{average maximum actuator work output} (AMAWO) as a function of the loading force for each frequency by computing \mbox{$\overline{W}_{\textrm{max}}(f_{\textrm{load}}) = f_{\textrm{load}} \cdot \bar{d}_{\textrm{max}}(f_{\textrm{load}})$}. For a frequency of $1$\,Hz, the best observed AMAWO value is of $1.4$\,{\textmu}J, which corresponds to a load of $1.26$\,mN. Clearly, for each tested load, the ALMADO value significantly decreases as the exciting frequency increases. Also, despite the existence of several outliers, for the first three frequencies, the data in Fig.\,\ref{FIG06}(b) indicates a decreasing trend of the ALMADO value as the load increases. Specifically, for the \mbox{frequency-DC} pair $\left\{1\,\textrm{Hz},6\,\textrm{\%}\right\}$, the mean of this figure of merit decreases from $1.625$ to $0.994$\,mm. Similarly, for the pair $\left\{5\,\textrm{Hz},11\,\textrm{\%}\right\}$, it decreases from $1.15$ to $0.655$\,mm; and, for the pair $\left\{10\,\textrm{Hz},10\,\textrm{\%}\right\}$, from $0.48$ to $0.315$\,mm. In contrast, for the pair $\left\{15\,\textrm{Hz},10\,\textrm{\%}\right\}$, the ALMADO value remains approximately constant as the load increases. It is not clear why this phenomenon occurs, but it might be related to the amount of kinetic energy in the system as a whole. Note that both the relatively large SEM values and output variations due to load increases highlight the need for feedback control in \mbox{real-time} applications of actuators of this type. Actuator failure typically occurs at a load of about $1.6$\,mN and, using simple microscopic analyses, we determined that this is caused by the fracture of the SMA material under mechanical stress. Also, the cause of failure provides empirical evidence of the structural integrity and high functionality of the proposed actuator. Consistently, its outstanding strength and work density are also evidenced by its ability to lift $155$ times its own weight for all the tested frequencies. To our best knowledge, no other microactuation technology compares to the presented method regarding work density. To further demonstrate these capabilities, in Section\,\ref{SECTION04}, we present two mobile microrobots driven by this actuation method. 
\begin{figure}[t!]
\vspace{1ex}
\begin{center}
\includegraphics{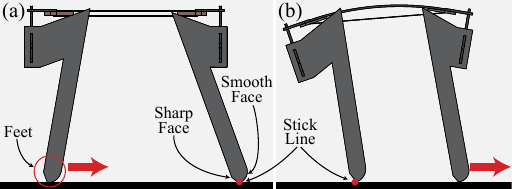}
\end{center}
\vspace{-2ex}
\caption{\textbf{Design and functionality of the MiniBug.} \textbf{(a)}~The MiniBug in its expanded state. The feet of this robot were designed with a sharp face and a smooth face to generate anisotropic friction and, as a consequence, forward locomotion during cyclic operation of the driving actuator. As the SMA wires of the driving actuator contract during heating, the sharp faces of the robot's front feet anchor to the supporting surface while the rear feet slide forward. \textbf{(b)}~The MiniBug in its contracted state. As the SMA wires of the driving actuator elongate during cooling, the sharp faces of the robot's rear feet anchor to the supporting surface while the front feet slide forward. \label{FIG07}}
\vspace{-2ex}
\end{figure}
\begin{figure*}[t!]
\vspace{1ex}
\begin{center}
\includegraphics{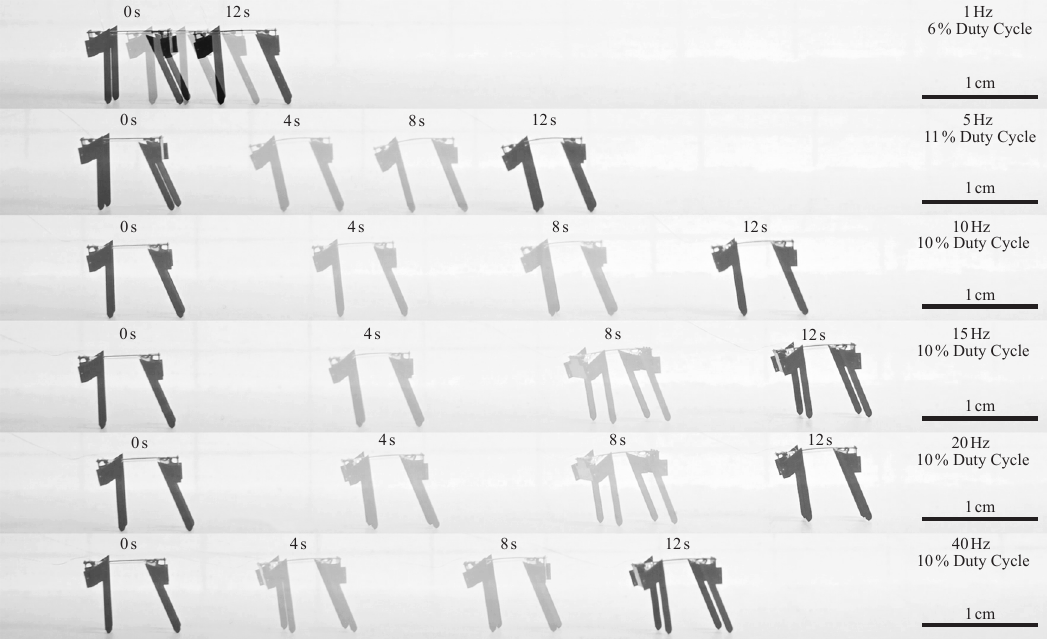}
\end{center}
\vspace{-2.0ex}
\caption{\textbf{The MiniBug locomoting at six different actuation frequencies.} The photographic sequences show the distance traveled by the MiniBug at intervals of $4\,\textrm{s}$, for the operation frequencies in the set $\left\{1,5,10,15,20,40\right\}$\,Hz; the DC values are the empirical optima determined according to the method discussed in~Section\,\ref{SECTION03}. The fastest relative speed of $0.76$\,BL/s occurs at $15$\,Hz. Depending on the frequency of operation, the robot exhibits four locomotion modes: (i)~\textit{crawling}, (ii)~\textit{shuffling}, (iii)~\textit{galloping}, and (iv)~\textit{gliding}. At $1$\,Hz, the MiniBug slowly crawls, achieving an average speed of $0.10$\,BL/s; at $5$\,Hz, the locomotion mode changes to shuffling, during which the MiniBug achieves an average speed of $0.46$\,BL/s; at $10$\,Hz, $15$\,Hz, and $20$\,Hz, the locomotion mode switches to galloping, during which the MiniBug achieves average speeds of $0.69$\,BL/s, $0.76$\,BL/s, and $0.75$\,BL/s, respectively. At the highest frequency of $40$\,Hz, the MiniBug \textit{glides} and hardly any actuator output displacement can be noticed as \mbox{high-frequency} vibrations allow the robot to \textit{virtually} float forward at an average speed of $0.61$\,BL/s. Video footage of these tests can be seen in the accompanying supplementary movie. \label{FIG08}}
\vspace{-1ex}
\end{figure*}
 
\section{Applications in Microrobotics}
\label{SECTION04}
\subsection{The MiniBug}
With a total weight of $8$\,mg and a length of $8.5$\,mm, the MiniBug (see Fig.\,\ref{FIG01}) is the lightest \mbox{fully-functional} \mbox{SMA-driven} crawler reported to date. This platform compellingly demonstrates the \mbox{high-frequency} capabilities of the proposed $0.96$-mg \mbox{SMA-based} actuator and its suitability to be integrated into microrobots. The MiniBug's base design was inspired by the SMALLBug presented in~\cite{SMALLBug}. However, a new \mbox{slot-and-pin} alignment method, along with a tuning procedure for its final physical configuration, enabled us to use $90$-{\textmu}m-thick CF material, which resulted in a significantly smaller and lighter robotic structure. We designed the feet of the MiniBug to cyclically and coordinately produce anisotropic friction and, as a consequence, forward locomotion. Each of the four feet has a sharp and a smooth face of contact with the supporting ground. Fig.\,\ref{FIG07} depicts the locomotion mechanisms during operation. Theoretically, as depicted in Fig.\,\ref{FIG07}(a), during heating, the actuator contracts and the front feet anchor to the ground as the stick lines of their sharp faces cling to the supporting surface while the back feet slide toward the right on their smooth faces. This contraction also moves the robot's \textit{center of mass} (COM) closer to the back legs, thus increasing the normal force acting on the back feet while anchored to the ground and facilitating the forward sliding of the front feet as the actuator expands during cooling, as shown in Fig.\,\ref{FIG07}(b). In reality, during crawling, the robot's feet slightly slide when they are supposed to be completely anchored; this phenomenon is more prominent at low locomotion frequencies ($1$\,Hz). This issue is mitigated at higher frequencies because the deflection drift of the driving actuator keeps the COM closer to the rear legs, which reduces undesired back sliding and, as a consequence, increases crawling efficiency.

To test and demonstrate the locomotion capabilities of the robot, we simply excited the driving actuator, in open loop, using a PWM signal with constant parameters during operation. As shown in Fig.\,\ref{FIG08}, we tested six cases corresponding to the frequency-DC pairs: $\left\{1\,\textrm{Hz},6\,\textrm{\%}\right\}$, $\left\{5\,\textrm{Hz},11\,\textrm{\%}\right\}$, $\left\{10\,\textrm{Hz},10\,\textrm{\%}\right\}$, $\left\{15\,\textrm{Hz},10\,\textrm{\%}\right\}$, $\left\{20\,\textrm{Hz},10\,\textrm{\%}\right\}$, and $\left\{40\,\textrm{Hz},10\,\textrm{\%}\right\}$. In all these tests, we kept the PWM \mbox{\textit{on}-voltage} height at $18$\,V, which we calculated using the total resistance of the actuator system in order to limit the resulting current to $200$\,mA. The photo sequences in Fig.\,\ref{FIG08} show $12$\,s of locomotion corresponding to the six experiments. Video footage of these tests can be seen in the accompanying supplementary movie. The MiniBug exhibits the same locomotion modes of the SMALLBug discussed in~\cite{SMALLBug}; however, at the highest actuation frequency of $40$\,Hz, a new locomotion mode is observed, which we dubbed as \textit{gliding}. In this mode, the driving actuator does not display noticeable actuator deflections; we speculate that the generated \mbox{high-frequency} vibrations induce the robot to slide forward. Because of the manner locomotion is produced at high frequencies, in this mode, traction is not significant and the MiniBug can be easily pushed around by minor disturbances. Everything considered, the best locomotion performance is achieved at the actuation frequency of $15$\,Hz because the robot simultaneously generates significant traction and locomotes at its fastest speed of $0.76$\,BL/s. In all tested modes, due to its small mass, the MiniBug is subjected to relatively large forces from the tether wires, which heavily affect locomotion speed. To remove these disturbances, we envision a tetherless power solution based on either \mbox{directed-energy} transmission, or catalytic combustion. 

\subsection{The WaterStrider}
\begin{figure}[t!]
\vspace{1.4ex}
\begin{center}
\includegraphics{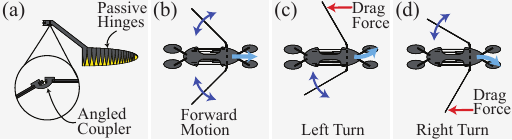}
\end{center}
\vspace{-1.2ex}
\caption{\textbf{Design and functionality of the WaterStrider.} \textbf{(a)}~The fin of the propulsor was designed to be flexible with a structure composed of a series of hinges, made of CF and Kapton, and fabricated using the SCM method. The robot was designed with its two propulsors biased towards the back of its body in order to reduce drag during forward locomotion. This characteristic was achieved using rigid \mbox{$30$-degree} angled couplers that connect, through CF bars, the transmissions and fins of the two propulsors. The couplers can be replaced to change the inclination angles of the propulsors. \textbf{(b)}~To generate forward thrust and, as a consequence, forward locomotion, the WaterStrider symmetrically flaps its two propulsors cyclically. \textbf{(c)}~To turn left, the WaterStrider flaps its right \mbox{fin-like} propulsor at a frequency of $5$\,Hz while its left propulsor remains inactive. The asymmetrical production of thrust plus the drag force acting on the inactive left propulsor generate a functional \mbox{counter-clockwise} torque on the robot's body. \textbf{(d)}~To turn right, the WaterStrider flaps its left \mbox{fin-like} propulsor at a frequency of $5$\,Hz while its right propulsor remains inactive. The asymmetrical production of thrust plus the drag force acting on the inactive right propulsor generate a functional clockwise torque on the robot's body. \label{FIG09}}
\vspace{-2ex}
\end{figure}

The extraordinary physical abilities exhibited by \mbox{water-surface-tension} locomoting insects have inspired scientific research~\cite{Bush2003Nature,Xuefeng2004Nature} and the development of new robots, such as those presented in~\cite{yan2015miniature}~and~\cite{koh2015jumping}. The robot in~\cite{yan2015miniature} weighs $1$\,g, stays afloat on hydrophobic wires, and uses sophisticated actuation mechanisms to generate elliptical stroke patterns for locomotion. However, its weight is much larger than that of \textit{Aquarius~paludum} specimens used for inspiration, which have a mass of only about $20$\,mg. The robot in~\cite{koh2015jumping} weighs $68$\,mg, stays afloat on hydrophobic wires, and can jump vertically $142$\,mm. In this case, SMA wires provide the actuation forces needed to jump off the water; however, this robot was not reported capable of locomotion. With a weight of $56$\,mg and a length of $22$\,mm, the WaterStrider (see Fig.\,\ref{FIG01}) is the lightest controllable \mbox{water-surface-tension} locomoting robot reported to date. We designed this robot to efficiently utilize the large force outputs produced by the unimorph \mbox{SMA-based} actuator presented in Section\,\ref{SECTION03}. The weight of the robot is supported by elliptical feet designed with large surface areas to exploit \mbox{surface-tension} forces while standing on water. We also designed the \mbox{highly-flexible} \mbox{fin-like} propulsor depicted in Fig.\,\ref{FIG09}(a) to take advantage of \mbox{fluid-structure-interaction} phenomena and thus increase hydrodynamic efficiency; actuating each propulsor independently enables locomotion and turning capabilities.

We used the modified SCM method in~\cite{yang202088} to fabricate all the structural and functional components of the WaterStrider, including two \mbox{four-bar} transmissions, four elliptical feet, a body frame, two \mbox{fin-like} propulsors (see Fig.\,\ref{FIG01}), and two \mbox{SMA-based} actuators of the type discussed in Section\,\ref{SECTION03}. Consistently, all these \mbox{multi-layer} parts were made from \mbox{$90$-{\textmu}m-thick} CF sheets and Kapton film. For the final assembly, we used the method presented in~\cite{yang202088} and the capillary alignment technique discussed in~Section\,\ref{SECTION02}. A key structural element of the WaterStrider's body frame are twisting reinforcement bars, installed to prevent \mbox{actuation-induced} body warping during operation. The main locomotion mode designed for the WaterStrider is depicted in~Figs.\,\ref{FIG09}(b)--(d). Here, the two \mbox{four-bar} transmission mechanisms receive as inputs the displacement outputs generated by the two \mbox{SMA-based} actuators and amplify them into large \mbox{stroke-angle} outputs that drive the \mbox{fin-like} bending propulsors of the system. As seen, during a locomotion period, the two actuators contract and then relax, enabling the propulsors to provide the force necessary to push the WaterStrider forward. During forward locomotion, both actuators are operated symmetrically, in open loop, to generate a large straight propulsion force. Accordingly, to execute turning maneuvers, one \mbox{actuator-propulsor} pair is actuated with a $5$-Hz PWM signal while the other \mbox{actuator-propulsor} pair is left inactive, which produces a \mbox{drag-based} turning torque on the robot. 
\begin{figure*}[t!]
\vspace{1ex}
\begin{center}
\includegraphics{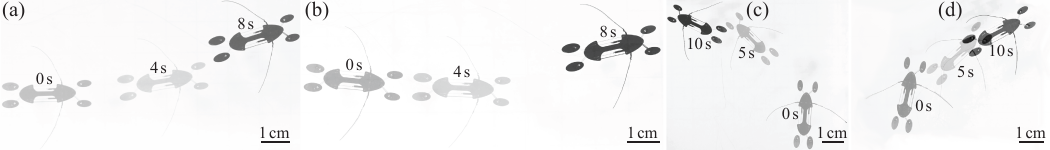}
\end{center}
\vspace{-2ex}
\caption{\textbf{The WaterStrider locomoting on water.} \textbf{(a)}~Photographic sequence showing the robot moving forward at an actuation PWM frequency of $1$\,Hz. The photographs in the composite were taken at $0$, $4$, and $8$\,s. \textbf{(b)}~Photographic sequence showing the robot moving forward at an actuation PWM frequency of $2$\,Hz. The photographs in the composite were taken at $0$, $4$, and $8$\,s. \textbf{(c)}~Photographic sequence showing a left turn. To execute this maneuver, the WaterStrider flaps its right \mbox{fin-like} propulsor at a PWM frequency of $5$\,Hz while its left propulsor remains inactive. The photographs in the composite were taken at $0$, $5$, and $10$\,s. \textbf{(d)}~Photographic sequence showing a right turn. To execute this maneuver, the WaterStrider flaps its left \mbox{fin-like} propulsor at a PWM frequency of $5$\,Hz while its right propulsor remains inactive. The photographs in the composite were taken at $0$, $5$, and $10$\,s. Video footage of these tests can be seen in the accompanying supplementary movie. \label{FIG10}}
\vspace{-2ex}
\end{figure*}

The photographic sequences in Fig.\,\ref{FIG10} summarize the experiments performed to test and demonstrate the locomotion capabilities of the WaterStrider. Video footage of these tests can be seen in the accompanying supplementary movie. The sequence in Fig.\,\ref{FIG10}(a) shows the WaterStrider during forward locomotion in open loop. In this case, both driving actuators were excited using $1$-Hz PWM signals with the optimal DC of $6$\,\%, empirically determined through the characterization experiments discussed in~Section\,\ref{SECTION03}. We selected an \mbox{\textit{on}-voltage} height of $12$\,V to ensure that no more than $200$\,mA of current passed through the SMA wires of the actuators; the same \mbox{\textit{on}-voltage} height was kept in all the experiments shown in~Fig.\,\ref{FIG10}. With these excitation parameters, the WaterStrider achieved an average speed of $0.26$\,BL/s. The sequence in Fig.\,\ref{FIG10}(b) also shows the WaterStrider during forward locomotion in open loop. In this case, however, both driving actuators were excited using $2$-Hz PWM signals with a DC of $7.5$\,\%. With these excitation parameters, the WaterStrider achieved an average speed of $0.28$\,BL/s.

To execute turning maneuvers, such as those shown in~Figs.\,\ref{FIG10}(c)--(d), one \mbox{actuator-propulsor} pair is \mbox{open-loop} excited using a $5$-Hz PWM signal with the optimal DC of $11$\,\%, determined through the characterization tests described in~Section\,\ref{SECTION03}, while the other \mbox{actuator-propulsor} pair is left inactive to produce a \mbox{drag-induced} body torque, as already explained. Specifically, in the test shown in Fig.\,\ref{FIG10}(c), the right \mbox{actuator-propulsor} pair is excited to make the robot turn left; similarly, in the test shown in Fig.\,\ref{FIG10}(d), the left \mbox{actuator-propulsor} pair is excited to make the robot turn right. During these left and right turning maneuvers, the WaterStrider achieved rates of $0.144$ and $0.073$\,rad/s, respectively. It is important to mention that the forces exerted by the tether wires on the WaterStrider during operation significantly affect its locomotion behavior because the friction induced by the water surface is almost negligible. For the same reason, the locomotion trajectory of its body can be easily and heavily disturbed by other external forces; this fact explains the observed difference in angular speed when the robot turns right and left. The ability of the WaterStrider to overcome these disturbances, while executing forward locomotion and turning maneuvers, demonstrates the capacity of the proposed \mbox{SMA-based} actuators to produce high output forces. Furthermore, these results indicate that once new onboard, or tetherless, technologies become available to power microrobots of the WaterStrider type, a wide gamut of applications useful for humans will become a reality. The most promising possibilities are \mbox{high-density} batteries~\cite{goldberg2018power}, catalytic combustion~\cite{yang202088}, and directed transmission of electromagnetic energy~\cite{kim2019laser}.

\section{Conclusions}
\label{SECTION05}
We presented a new \mbox{$0.96$-mg} (\mbox{$\sim\hspace{-0.6ex}1$\,mg}) fast unimorph \mbox{SMA-based} actuator that is capable of \mbox{high-frequency} operation (up to $40$\,Hz) as well as lifting $155$ times its own weight. This development is the result of using the modified SCM method in~\cite{beePlus2019,yang202088} and the introduction of a new alignment technique for microfabrication based on the use of passive capillary forces. Through dynamic characterization experiments, we tested and demonstrated the \mbox{high-frequency} operation and \mbox{high-force} output capabilities of the proposed \mbox{SMA-based} actuator. To show the suitability of the actuator in microrobotic applications, we designed and built two locomoting microrobots: (i)~the MiniBug, which, with a weight of $8$\,mg and a length of $8.5$\,mm, is the lightest \mbox{fully-functional} \mbox{SMA-driven} terrestrial crawler reported to date; and, (ii)~the \mbox{$56$-mg} \mbox{$22$-mm--long} WaterStrider, which is the first \mbox{subgram} controllable \mbox{SMA-driven} crawler capable of locomoting on water by taking advantage of \mbox{surface-tension} effects. Through the discussion of several tests, we demonstrated the locomotion behavior and performance of the MiniBug during operation, which can function at actuation frequencies of up to $40$\,Hz and achieve an average speed of $0.76$\,BL/s. Similarly, we demonstrated the locomotion behavior and performance of the WaterStrider during operation, which can achieve an average speed of $0.28$\,BL/s and execute turning maneuvers at angular rates of up to $0.144$\,rad/s. To achieve autonomy, we envision the deployment of MiniBug and WaterStrider platforms in swarms that collectively would be capable of carrying enough power to complete missions. 
\bibliographystyle{IEEEtran}
\bibliography{references}
\end{document}